\documentclass[11pt,a4paper]{article}
\usepackage[hyperref]{naaclhlt2019}
\usepackage{times}
\usepackage{latexsym}

\usepackage{colortbl}
\usepackage{amsmath}

\aclfinalcopy

\title{Language and Dialect Identification of Cuneiform Texts}

\author{Tommi Jauhiainen \\
  Department of Digital Humanities \\
  University of Helsinki \\
  {\tt tommi.jauhiainen@helsinki.fi} \\\And
  Heidi Jauhiainen \\
  Department of Digital Humanities \\
  University of Helsinki \\
  {\tt heidi.jauhiainen@helsinki.fi} \\\AND
  Tero Alstola \\
  Department of Digital Humanities \\
  University of Helsinki \\
  {\tt tero.alstola@helsinki.fi} \\\And
  Krister Lind\'en \\
  Department of Digital Humanities \\
  University of Helsinki \\
  {\tt krister.linden@helsinki.fi} \\}

\date{}

\begin{document}
\maketitle
\begin{abstract}

This article introduces a corpus of cuneiform texts from which the dataset for the use of the Cuneiform Language Identification (CLI) 2019 shared task was derived as well as some preliminary language identification experiments conducted using that corpus. We also describe the CLI dataset and how it was derived from the corpus. In addition, we provide some baseline language identification results using the CLI dataset. To the best of our knowledge, the experiments detailed here represent the first time that automatic language identification methods have been used on cuneiform data.

\end{abstract}

\section{Introduction}

We have compiled a corpus of cuneiform texts intended to be used in language identification experiments. As the basis for our corpus, we used the Open Richly Annotated Cuneiform Corpus (Oracc).\footnote{[http://oracc.museum.upenn.edu]} In Oracc, the texts are stored in transliterated form. We created a tool, Nuolenna, which can transform the transliterations back to the cuneiform script. Selecting all monolingual lines from Oracc and transforming the transliterations into cuneiform, we created a new corpus for Sumerian and six Akkadian dialects.

This corpus was used in the initial experiments where the possibility of language identification in cuneiform texts was verified. In this paper, we report some of the results from the initial experiments. To the best of our knowledge, this is the first time that automatic language identification methods have been used on cuneiform data. The methods we use for language identification utilize mainly character \emph{n}-grams and their observed probabilities in text.

For the use of the Cuneiform Language Identification (CLI) 2019 shared task\footnote{[https://sites.google.com/view/vardial2019/campaign]}, we extracted a dataset from the corpus.
The dataset is divided into training, development, and test portions to be used in the CLI shared task which is part of the third VarDial Evaluation Campaign. We implemented four baseline language identifiers and evaluated their performance using the CLI dataset. The results of the evaluation are presented here.

\section{Related work}

So far, no research into language identification using cuneiform texts has been openly reported. Language identification studies involving other contemporary scripts, such as Egyptian hieroglyphs, also seem to be non-existent.

\subsection{Cuneiform script and computational methods}

In this section, we survey some of the research where computational methods related to language identification have been used with the cuneiform script.

\citet{kataja1} discuss the description and computational implementation of phonology and morphology for Akkadian. They give examples of the rules in two-level formalism they used with the TWOL rule compiler.

\citet{barthelemy1} developed and tested a morphological analyzer for Akkadian verbal forms. The analyzer works with Akkadian represented in Latin encoding (transcription).

\citet{tablan1} describe their project, which aims to create a tool for automatic morphological analysis of Sumerian.

Among several languages, \citet{rao1} analyzed Sumerian written in cuneiform using conditional entropy to compare it with the Indus script. Normalized entropy of sign \emph{n-grams} between the two scripts was used as further evidence to indicate the possible linguistic content of the texts written in the Indus script.

\citet{ponti1} used the K-means clustering algorithm to cluster transliterated cuneiform texts. The texts analyzed were 51 Old Babylonian letters from Tell Harmal/\v{S}adupp\^um. \textit{Term frequency} (TF) and \textit{term frequency - inverse document frequency} (TF-IDF) weighted words were used as features with the clustering methods. Each document was depicted as a feature vector with the length of the whole vocabulary. In K-means, the number of clusters has to be given before the algorithm is applied and \citet{ponti1} experimented with 2 to 15 clusters.

\citet{luo1} describe an unsupervised Named-Entity Recognition (NER) system for transliterated Sumerian. They compared the use of different lengths of transliterated word \emph{n}-grams in combination with the Decision List CoTrain algorithm, and their evaluations show that word bigrams obtain the highest F1-score. In another article \cite{liu1}, they describe how they managed to find unannotated personal names in a corpus and suggest that the NER system could be used as an automated tool for the annotation task. \citet{liu2} continue the NER research on Sumerian using a variety of supervised classification methods to detect named entities.

\citet{homburg1} researched automated word segmentation of Akkadian cuneiform script. They used a sign list to restore CDLI\footnote{Cuneiform Digital Library Initiative [https://cdli.ucla.edu]} transliterations back to cuneiform (represented as UTF-8 characters). This is the only related work we are aware of in which cuneiform texts encoded in Unicode cuneiform have been processed previous to our experiments.

\citet{pageperron1} present a project dedicated to creating a pipeline for Sumerian texts. The pipeline is planned to take in transliterated Sumerian and to produce POS tag annotations and lemmatization as well as machine translation into English. \citet{chiarcos1} further describe the work done in the project.

In order to measure inter-textual relations, \citet{monroe1} calculated the cosine similarity between word vectors consisting of transliterated Late Babylonian words.

\citet{svard1} used Pointwise Mutual Information (PMI) to find collocations and associations between words and word2vec to highlight paradigmatic relationships of the words of interest. They used transliterated and lemmatized Akkadian texts from Oracc.

\subsection{Language identification}\label{LI}

Automatic language identification is the task of determining the language of a piece of text from the clues in the text itself. The computational methods used in language identification vary from simple wordlists to state-of-the-art deep learning methods. A recent comprehensive survey on language identification was conducted by \citet{jauhiainen1}. Language identification for long texts in well-resourced languages is not a difficult task, but it becomes increasingly more challenging when we target short, fragmentary, and multilingual texts in languages where the amount of training material is severely restricted. A separate challenge for language identification is dealing with closely related languages or with several dialects of an individual language. The challenge of discriminating between closely related languages has been investigated in a series of shared tasks that have been organized as part of VarDial workshops \citep{zampieri6,zampieri8,malmasi9,zampieri9,zampieri12}.

\section{Cuneiform texts in Oracc}

Our data comes from the Open Richly Annotated Cuneiform Corpus (Oracc). Oracc is an international cooperative effort containing free online editions of cuneiform texts from various projects.\footnote{[http://oracc.museum.upenn.edu/doc/about/aboutoracc/ index.html]} Oracc is one of the largest electronic corpora of Sumerian and Akkadian texts, and it is regularly updated. Our data is a snapshot of Oracc from October 2016 from XML files downloaded with the permission of the site administrators. The data is comprised of 13,662 separate texts, most of which were originally written on clay tablets. Some of the texts are duplicates, and before the language identification we removed the duplicates of texts with identical Oracc identification numbers. This procedure removes modern duplicates which have come into existence because a single text has been edited in several Oracc projects. Those duplicates that have different numbers and are thus different ancient manuscripts were not removed. Cuneiform writing does not mark the end of a sentence, and this is not indicated in the XML files either. Our data can, hence, be divided either into lines or texts with one or more lines. Oracc also contains some texts or words written in languages other than Sumerian and Akkadian, such as Hittite, Ugaritic, and Greek, but their number is so small that they were left out of this research.

The metadata of the texts in Oracc contains information about, for example, the provenance (the locality where the text was found), the genre, the time period in which the text was written, and so forth. The basic units in Oracc XML files are the transliterations of words, which are representations of the cuneiform signs in Latin script and which are given even if nothing else is stated about the words. Some of the cuneiform signs have, however, been broken off or are otherwise unreadable on the original tablets. In those cases, the word in question, or part of it, is replaced with an 'x' in the transliteration. The metadata of a word usually indicates its language, and some of the projects have also provided the cuneiform signs for each syllable or word of the transliteration.

The data contains many bilingual documents written in Sumerian and Akkadian. These documents often have the same text in both languages, sometimes on the same line.

\subsection{Sumerian and Akkadian}
Sumerian and Akkadian are ancient languages which were spoken and written primarily in Mesopotamia, present-day Iraq \cite {Michalowski1, Kouwenberg1}. Both languages were written in cuneiform script, but they are not related, Sumerian being a language isolate and Akkadian an East Semitic language. The cuneiform script was originally logographic in essence, then syllabic sign values were introduced to facilitate writing Sumerian, and only later was the script adapted for Akkadian. Consequently, some features of the cuneiform writing system are not ideal for Akkadian and many logograms are used side by side with syllabic spellings of Akkadian words \cite[for further information see][]{seri10}.

Sumerian was one of the first languages ever written, and the oldest texts survive from the turn of the fourth and third millennia before the Common Era (BCE). Akkadian replaced Sumerian as the spoken language during the late third and early second millennia BCE, but Sumerian was used as a liturgical and scholarly language until the end of the cuneiform tradition at the beginning of the Common Era.

Written Akkadian is known from circa 2400 BCE onwards until the first century CE. The Akkadian language had two main dialects, Assyrian and Babylonian, both of which are present in our data. Assyrian was used in northern Mesopotamia and Babylonian in the south. There is written evidence for the simultaneous use of these dialects for 1,400 years, and both of them changed over time. The dialects are, hence, further divided into varieties designated as Old, Middle, and Neo-Assyrian and Old, Middle, and Neo-Babylonian. There was also a literary variety called Standard Babylonian which was used by both Assyrian and Babylonian scribes to write texts in certain genres. In Oracc, Middle-Babylonian is, furthermore, divided into the dialect spoken in Mesopotamia proper and the one spoken in the peripheries of the Empire. The latter, referred to as Middle Babylonian Peripheral, is not a coherent dialect but varies somewhat from site to site. After Assyrian ceased to be a written language around 600 BCE, a variety called Late Babylonian was written for some 700 years. The differences between the dialects and their varieties are relatively small, and after learning a variety one can read the other dialect and varieties as well. In the Oracc metadata, the different dialects and varieties are given for Akkadian words in most cases.

\section{Cuneiform representation in Unicode}

The effort to provide a standard encoding for cuneiform began in 1999 at Johns Hopkins University as the Initiative for Cuneiform Encoding (ICE). The initiative ended up with an approved proposal for cuneiform Unicode in 2004 (officially accepted into Unicode 5.0 in 2006).\footnote{The Unicode® Standard Version 11.0 – Core Specification [http://www.unicode.org/versions/Unicode11.0.0/appC.pdf]} The final list of cuneiform signs included is a combination of work done earlier at the University of Chicago, Universit{\"a}t G{\"o}ttingen, and the University of California, Los Angeles \cite{cohen1}. 

In the current Unicode standard, there are three blocks of cuneiform signs for the ``Sumero-Akkadian'' script. The first one is the block covering the base cuneiform signs ranging from U+12000 to U+123FF. The second block, from U+12400 to U+1247F, covers the cuneiform punctuation and numerals and the third, from U+12480 to U+1254F, is an extension containing additional signs for the Early Dynastic period. Unicode has only one character for each sign, even though the signs evolved through the ages. The different ways of writing the signs could be used to determine the language or dialect used or the time period of writing.

The cuneiform texts from the Oracc corpus we use in this research were provided primarily as transliterations using the ASCII Transliteration Format (ATF). ATF was first defined by CDLI and is a standardized way of electronically transliterating cuneiform, following the conventions used by cuneiform scholars in general \cite{koslova1}.  The data from the Oracc corpus is also available as JSON files in an ``XCL'' format \cite{tinney1}\footnote{[http://oracc.museum.upenn.edu/doc/opendata/]}, which includes a similar XML representation of the data as the CDLI archival XML format~\cite{koslova1}. We extracted the individual sign transliterations in Unicode ATF from the XML representation and recreated the transliteration for each line.

There was no available software to automatically transform the transliterations to Unicode cuneiform. As part of Oracc, there is a facility called ``Cuneify,'' which can be used online to transform ATF into Unicode cuneiform.\footnote{[http://oracc.museum.upenn.edu/doc/tools/cuneify/ index.html]} However, it is not possible to download the software and it does not handle the Unicode ATF transliteration. In order to generate the original lines in cuneiform, we implemented a program called ``Nuolenna'' which takes in the transliteration generated from the XML files and re-produces the lines in Unicode cuneiform.\footnote{[https://github.com/tosaja/Nuolenna]} The Nuolenna program uses a list\footnote{[https://github.com/tosaja/Nuolenna/blob/master/ sign\_list.txt]} of over 11,000 transliteration-sign pairs. As the base for our sign list, we used a JSON export from the Oracc Global Sign List (OGSL)\footnote{[http://oracc.museum.upenn.edu/ogsl/]} provided by Niek Veldhuis, to which we added some missing signs. In order to produce the original cuneiform lines, the program uses \textit{ad hoc} rules to remove any additional annotations related to the signs. For example, sometimes an older or more precise reading can be found within parentheses directly after the reading of a sign. In such cases, we just remove the parentheses and anything between them.

\section{Preliminary language identification experiments}

To find out to what extent identifying the language of cuneiform text is possible, we performed initial language identification experiments using a state-of-the-art language identification method called HeLI \citep{jauhiainen2}. The HeLI method has recently fared well in VarDial shared tasks for Swiss-German dialect and Indo-Aryan language identification \citep{jauhiainen3,jauhiainen4}. The experiments were conducted on individual lines as well as texts spanning several lines.

\subsection{Corpus for the preliminary experiments}

In Oracc, the transliterated words are separated by whitespaces, which is not the case in the original documents. In order to mimic the original documents, we removed all the whitespaces from each line of cuneiform text. We also ignored any completely broken signs, which were marked with an 'x'.

The individual words in Oracc are tagged with language or dialect information, and sometimes a single line includes words in different languages or dialects. As we set out to do language identification on monolingual texts, we used all those lines which had words in only one language, leaving out multilingual lines. The language tagging in Oracc is not always precise, and therefore some lines in our dataset might still include several languages.

In the preliminary experiments, we experimented with the language identification of both monolingual lines and monolingual texts spanning several lines with the information about line breaks retained. Mostly, each text had the lines of one original document, but if the document was multilingual, it was divided into different texts according to the languages attested.

We left out the Akkadian varieties of Old and Middle Assyrian as the number of lines available for those dialects was less than 1,000. We had datasets in the Sumerian language as well as the Akkadian varieties of Old Babylonian, Middle Babylonian peripheral, Standard Babylonian, Neo-Babylonian, Late Babylonian, and Neo-Assyrian. The statistics of the corpus used in the preliminary experiments are shown in Table~\ref{corpusstats1}.

\begin{table*}[t!]
\begin{center}
\footnotesize
\begin{tabular}{|l|rrl|}
\hline 
\rowcolor[gray]{.8} \bf Language or Dialect (abbreviation in the CLI dataset) & \bf Texts & \bf Lines & \bf Signs \\ \hline
Sumerian (SUX) & 5,000 & 107,345 & \emph{c}. 400,000\\
\rowcolor[gray]{.9} Old Babylonian (OLB) & 527 & 7,605 & \emph{c}. 65,000\\
Middle Babylonian peripheral (MPB) & 365 & 11,015 & \emph{c}. 95,000\\
\rowcolor[gray]{.9} Standard Babylonian (STB) & 1,661 & 35,633 & \emph{c}. 390,000 \\
Neo-Babylonian (NEB) & 1,212 & 19,414 & \emph{c}. 200,000 \\
\rowcolor[gray]{.9} Late Babylonian (LTB) & 671 & 31,893 & \emph{c}. 260,000 \\
Neo-Assyrian (NEA) & 3,570 & 65,932 & \emph{c}. 490,000\\
\hline
\end{tabular}
\end{center}
\caption{\label{corpusstats1} Number of texts, lines, and signs for each language or variety in the corpus.}
\end{table*}

We were interested in experimenting in both in-domain and out-of-domain test settings as well as with language identification on two different levels: individual lines and texts. In supervised machine learning, the testing data is in-domain if it is similar to the training data. For example, if sentences are from texts that belong to the same genre or collection they are considered more in-domain than if they are not. An even stronger in-domain case is if the sentences are from the same text. Classification of test data which is in-domain with the training data is usually much easier than when it is out-of-domain. The texts in the Oracc export were in the order\footnote{The projects were in the alphabetical order by their abbreviations.} of ``projects,'' which are collections of texts that have some common theme. The texts in different projects can be considered to be more out-of-domain with each other than those from the same project. The projects from which the texts were extracted are listed in Table~\ref{projects1}.

\begin{table*}[t!]
\begin{center}
\footnotesize
\renewcommand{\tabcolsep}{1mm}
\begin{tabular}{|l|ccccccc|}
\hline 
\rowcolor[gray]{.8} \bf Project (abbreviation used in Oracc) & \bf SUX & \bf OLB & \bf MPB & \bf STB & \bf NEB & \bf LTB & \bf NEA \\ \hline
Bilinguals in Late Mesopotamian Scholarship (\textbf{blms}) & x & x & & x & & & \\
\rowcolor[gray]{.9} CAMS/Anzu (\textbf{cams-anzu}) & & & & x & & & \\
CAMS/Barutu (\textbf{cams-barutu}) & & x & & x & & & \\
\rowcolor[gray]{.9} CAMS/The Standard Babylonian Epic of Etana (\textbf{cams-etana}) & & x & & & & & x \\
CAMS/Geography of Knowledge Corpus (\textbf{cams-gkab}) & x & & & x & & x & x \\
\rowcolor[gray]{.9} CAMS/Ludlul (\textbf{cams-ludlul}) & & & & x & & & \\
CAMS/Seleucid Building Inscriptions (\textbf{cams-selbi}) & & & & x & & & \\
\rowcolor[gray]{.9} Cuneiform Commentaries Project on ORACC (\textbf{ccpo}) & x & & & x & & & \\
Corpus of Kassite Sumerian Texts (\textbf{ckst}) & x & & & & x & & \\
\rowcolor[gray]{.9} The Amarna Texts (\textbf{contrib-amarna}) & & & x & x & & & \\
Cuneiform Texts Mentioning Israelites, Judeans ... (\textbf{ctij}) & & & & & x & x & x \\
\rowcolor[gray]{.9} Lexical Texts in the Royal Libraries at Nineveh (\textbf{dcclt-nineveh}) & x & & & x & & & \\
Reading the Signs (\textbf{dcclt-signlists}) & x & & & x & & & \\
\rowcolor[gray]{.9} Digital Corpus of Cuneiform Lexical Texts (\textbf{dcclt}) & x & x & x & x & & & \\
Digital Corpus of Cuneiform Mathematical Texts (\textbf{dccmt}) & x & x & & x & & & \\
\rowcolor[gray]{.9} Electronic Text Corpus of Sumerian Royal Inscriptions (\textbf{etcsri}) & x & & & & & & \\
Corpus of Glass Technological Texts (\textbf{glass}) & & & & x & & & \\
\rowcolor[gray]{.9} Hellenistic Babylonia: Texts, Iconography, Names (\textbf{hbtin}) & & & & & & x & \\
Law and Order: Cuneiform Online Sustainable Tool (\textbf{lacost}) & x & & & & & & \\
\rowcolor[gray]{.9} Old Babylonian Model Contracts (\textbf{obmc}) & x & & & & & & \\
Old Babylonian Tabular Accounts (\textbf{obta}) & x & x & & & & & \\
\rowcolor[gray]{.9} The Inscr. of the Second Dynasty of Isin (\textbf{ribo-babylon2}) & x & & & & & & \\
The Inscr. of the Period of the Uncertain Dynasties (\textbf{ribo-babylon6}) & x & & & & & & \\
\rowcolor[gray]{.9} Rim-Anum: The House of Prisoners (\textbf{rimanum}) & x & x & & & & & \\
The Correspondence of Sargon II, Part I (\textbf{saao-saa01}) & & & & & & & x \\
\rowcolor[gray]{.9} Neo-Assyrian Treaties and Loyalty Oaths (\textbf{saao-saa02}) & & & & x & x & & x \\
Court Poetry and Literary Miscellanea (\textbf{saao-saa03}) & & & & & x & & x \\
\rowcolor[gray]{.9} Queries to the Sungod (\textbf{saao-saa04}) & & & & & x & & \\
The Correspondence of Sargon II, Part II (\textbf{saao-saa05}) & & & & & & & x \\
\rowcolor[gray]{.9} Legal Trns. of the Royal Court of Nineveh, Part I (\textbf{saao-saa06}) & & & & & & & x \\
Imperial Administrative Records, Part I (\textbf{saao-saa07}) & & & & & & & x \\
\rowcolor[gray]{.9} Astrological Reports to Assyrian Kings (\textbf{saao-saa08}) & & & & x & x & & x \\
Assyrian Prophecies (\textbf{saao-saa09}) & & & & & & & x \\
\rowcolor[gray]{.9} Letters from Assyrian and Babylonian Scholars (\textbf{saao-saa10}) & & & & x & x & & x \\
Imperial Administrative Records, Part II (\textbf{saao-saa11}) & & & & & & & x \\
\rowcolor[gray]{.9} \rowcolor[gray]{.9} Grants, Decres and Gifts of the Neo-Assyrian Period (\textbf{saao-saa12}) & & & & & & & x \\
Letters from Assyrian and Babylonian Priests to ... (\textbf{saao-saa13}) & & & & & x & & x \\
\rowcolor[gray]{.9} Legal Trns. of the Royal Court of Nineveh, Part II (\textbf{saao-saa14}) & & & & & & & x \\
The Correspondence of Sargon II, Part III (\textbf{saao-saa15}) & & & & & & & x \\
\rowcolor[gray]{.9} The Political Correspondence of Esarhaddon (\textbf{saao-saa16}) & & & & & & & x \\
The Neo-Babylonian Correspondence of Sargon and ... (\textbf{saao-saa17}) & & & & & x & & \\
\rowcolor[gray]{.9} The Babylonian Correspondence of Esarhaddon and ... (\textbf{saao-saa18}) & & & & & x & & \\
The Correspondence of Tiglath-Pileser III and ... (\textbf{saao-saa19}) & & & & & x & & x \\
\hline
\end{tabular}
\end{center}
\caption{\label{projects1} The list of Oracc projects from which the texts in the corpus were collected.}
\end{table*}

From this corpus, we generated four different test settings. For the out-of-domain experiments, we divided the corpus so that we used the first half of the corpus for training and the second half was divided between development and testing. For the in-domain experiments, we divided the corpus into parts of 20 lines or texts and took the 10 first lines or texts from each part for training, the next 5 for development, and the last 5 for testing. We, thus, ended up with four different datasets,\footnote{See Table~\ref{heli1}.} two for lines and two for texts. Each of the datasets had 50\% of the material for training, 25\% for development, and 25\% for testing.

\subsection{Results of the preliminary experiments}

The HeLI method is a supervised-learning language identification method where the language models are created from a correctly tagged training corpus. The language models consist of words and sign (character) \emph{n}-grams. When \emph{n}-grams are extracted from a corpus, the number of unique \emph{n}-grams is higher the longer the \emph{n}-grams are. The actual number of occurrences of the longer \emph{n}-grams is lower than the shorter \emph{n}-grams. The exact optimal value for \emph{n} depends on, among many other variables, the size of the training corpus, the length of the text to be identified, and the repertoire of the languages considered. Sometimes the longer \emph{n}-grams could carry important information even though they are seldom found in the text to be identified. The basic idea in the HeLI method is to score individual words using the longest length \emph{n}-grams possible. For each individual language, the words are scored first, after which the whole text gets the average score of the individual words. In the case of cuneiform text, as it is not divided into words, we use just sign \emph{n}-grams and consider a line of text as a word as far as the HeLI method is concerned.

The individual words, or in this case lines, are scored by taking the average score of the found \emph{n}-grams. Using the notation introduced by \citet{jauhiainen1}, the individual \emph{n}-grams \(f\) found from the line to be tested, get a score \(R\) as in Equation~\ref{eqn:relativefrequency}:

\begin{equation}
\footnotesize
R_{HeLI}(g,f)=-\log_{10} \frac {c(C_g,f)} {l_{C_g^{F}}}
\label{eqn:relativefrequency}
\end{equation}
where \(c(C_g,f)\) is the count of the feature \(f\) in the training corpus \(C_g\) of the language \(g\) and \(l_{C_g^{F}}\) is the total number of occurrences of all the \emph{n}-grams of the same length in the training corpus. As smoothing, in case the count of a feature is zero in some languages, this version of the HeLI method uses a score \(R_{HeLI}(g,f)\) for the count of one multiplied by a penalty multiplier.

Using the development sets, we optimized the sign \emph{n}-gram range and the penalty multiplier for each setting individually. The results of these experiments are presented in Table~\ref{heli1}. As the performance measure, we use the F1-score which is the harmonic mean of precision and recall. The results clearly show how the task of identifying a single line is much harder than that of a complete text. The task of out-of-domain identification is also clearly more difficult than that of in-domain, as was expected.

\begin{table}[t!]
\begin{center}
\footnotesize
\begin{tabular}{|l|r|l|}
\hline 
\rowcolor[gray]{.8} \bf Type of setting & \bf \emph{n}-gram range & \bf F1 \\ \hline
Lines, out-of-domain & 1--3 & 60\\
\rowcolor[gray]{.9} Lines, in-domain & 1--3 & 72\\
Texts, out-of-domain & 1--4 & 84\\
\rowcolor[gray]{.9} Texts, in-domain & 1--3 & 93\\
\hline
\end{tabular}
\end{center}
\caption{\label{heli1} The F1-scores attained by the HeLI method in the preliminary experiments.}
\end{table}

Quite many of the misclassified lines were very short; many consisted only of one sign and were truly ambiguous and often present in different dialects and even languages. Nevertheless, it was still possible to attain reasonably good language identification results. The hardest test setting was where the language of individual out-of-domain lines was to be identified. To us, this seemed to be the most interesting and relevant setting to be used in the CLI shared task, especially if we leave out the extremely short and possibly ambiguous lines.

\section{The CLI shared task}

The CLI shared task 2019, part of the third VarDial Evaluation Campaign, focused on discriminating between languages and dialects written with cuneiform signs. The task included two different languages: Sumerian and Akkadian. Furthermore, the Akkadian language was divided into six dialects: Old Babylonian, Middle Babylonian peripheral, Standard Babylonian, Neo-Babylonian, Late Babylonian, and Neo-Assyrian. First, we explain how the dataset for the shared task was constructed from the corpus described earlier, and then we present the baseline language identifiers and the results we attained using them.

\subsection{The dataset for the shared task}

For the CLI task, we created a separate, especially tailored dataset. The participants were given texts for training and development and separate texts were given for testing at the end of the campaign. The training set was exactly the same as the one we used in the preliminary experiments\footnote{In the out-of-domain individual line identification test setting.} and the number of lines in the training portion for each language or dialect is shown in Table~\ref{corpusstats2}.

\begin{table}[t!]
\begin{center}
\footnotesize
\begin{tabular}{|l|r|}
\hline 
\rowcolor[gray]{.8} \bf Language or Dialect & \bf Training \\ \hline
Sumerian & 53,673 \\
\rowcolor[gray]{.9} Old Babylonian & 3,803 \\
Middle Babylonian peripheral & 5,508 \\
\rowcolor[gray]{.9} Standard Babylonian & 17,817 \\
Neo-Babylonian & 9,707 \\
\rowcolor[gray]{.9} Late Babylonian & 15,947\\
Neo-Assyrian & 32,966\\
\hline
\end{tabular}
\end{center}
\caption{\label{corpusstats2} Number of lines for each language or dialect in the training set provided during the VarDial 2019 Evaluation Campaign.}
\end{table}

For the CLI development and test sets, we performed some further operations. The original datasets included duplicate lines, so we first removed all duplicates. Then we filtered out all lines shorter than three characters. After these operations, the smallest sets were those of Old Babylonian including 668 lines in the development set and 985 lines in the test set. As we wanted to make the development and the test sets equal in size between languages and dialects, we randomly selected the same number of lines from the other languages. Thus, in the CLI task, the development sets for each language consisted of 668 lines and the test sets of 985 lines.

\subsection{Baseline experiments}

We used four of the methods described in the survey by \citet{jauhiainen1} to implement baseline language identifiers for the CLI task. As features, we used sign \emph{n}-grams of different lengths.

The first method is called simple scoring. In simple scoring, all the \emph{n}-grams generated from the line to be identified \(M\) are compared to the language models and for each \emph{n}-gram found in a language model \(dom(O(C_{g}))\), the score \(R\) of the language \(g\) is increased by one. The language gaining the highest score is selected as the predicted language. \citet{jauhiainen1} formulate the method as in Equation~\ref{eqn:simple1}:

\begin{equation}
\footnotesize
\label{eqn:simple1}
R_{\text{\emph{simple}}}(g,M) = \sum_{i=1}^{l_{M^F}} \left\{
  \begin{array}{ll}
    1 & \textbf{, if } f_{i} \in dom(O(C_{g}))\\
    0 & \textbf{, otherwise}
  \end{array} \right.
\end{equation}
where \(l_{M^F}\) is the number of individual features in the line \(M\) and \(f_{i}\) is its \(i\)th feature.

The second method is the sum of relative frequencies where relative frequencies are added to the score of the language. \citet{jauhiainen1} formulate the method as in Equation~\ref{eqn:sumvalues1}:

\begin{equation}
\label{eqn:sumvalues1}
\footnotesize
R_{\text{\emph{sum}}}(g,M)= \sum_{i=1}^{l_{M^F}} \frac {c(C_g,f_i)} {l_{C_g^{F}}}
\end{equation}
where \(c(C_g,f_i)\) is the count of the feature \(f_i\) in the training corpus. 

The third method is the product of relative frequencies where the relative frequencies are multiplied together. \citet{jauhiainen1} formulate the method as in Equation~\ref{eqn:productvalues1}:

\begin{equation}
\label{eqn:productvalues1}
\footnotesize
R_{\text{\emph{prod}}}(g,M)= \prod_{i}^{l_{M^F}} \frac {c(C_g,f_i)} {l_{C_g^{F}}}
\end{equation}

The actual implementation adds together negative logarithms of the relative frequencies, which produces results with the same ordering. As a smoothing value, we used the negative logarithm of a comparably small relative frequency. The actual value was optimized using the development set.

The fourth method is a majority-voting-based ensemble of the three previous methods. 

The parameters and the best possible language models are determined by training the identifier using the training set and evaluating its performance on the development set. Once the best parameters are decided, the texts in the development set can also be added to the training set for the final evaluation against the test set. We used the macro F1-score as the measure for language identification performance. For each of the methods, we evaluated all possible sign \emph{n}-gram ranges from 1 to 15 using the development set. Table~\ref{cli1} shows the results for all the methods using parameters optimized with the development set. In the voting ensemble, we used the best parameters for the methods from the individual experiments, and in case of a tie, the result from the product of relative frequencies was used.

\begin{table}[t!]
\begin{center}
\footnotesize
\begin{tabular}{|l|c|c|c|}
\hline 
\rowcolor[gray]{.8} \bf Method & \bf \emph{n}-gram range & \bf F1 dev & \bf F1 test \\ \hline
Prod. of rel. freq. & 1--4 & 0.7263 & 0.7206\\
\rowcolor[gray]{.9} Voting Ensemble & & 0.7222 & 0.7163\\
HeLI & 1--3 + lines & 0.7171 & 0.7061\\
\rowcolor[gray]{.9} Simple scoring & 1--10 & 0.6656 & 0.6554\\
Sum of rel. freq. & 3--15 & 0.5984 & 0.6016\\
\hline
\end{tabular}
\end{center}
\caption{\label{cli1} The macro F1-scores attained by the baseline methods with the CLI dataset.}
\end{table}

The product of relative frequencies method is clearly superior to the other two basic methods with an F1-score of 0.7206 using 2.0 as the smoothing value and sign \emph{n}-grams from one to four. Adding the prediction information from the other two methods in the form of a voting ensemble also fails to improve the result. The F1-score achieved when using the HeLI method does not reach the one from the product of relative frequencies method either. The F1-score gained by the HeLI method is clearly higher than the score attained in the preliminary experiments, which was as expected, as we had filtered out some of the most difficult cases.

Table~\ref{cli1} is a confusion matrix displaying the exact number of identifications. The diagonal values represent correct identifications. Standard Babylonian and Neo-Babylonian were the most difficult varieties to distinguish, mostly being erroneously identified as each other. Late Babylonian was the easiest to identify, with a recall of over 96\%.

\begin{table}[t!]
\begin{center}
\footnotesize
\renewcommand{\tabcolsep}{1mm}
\begin{tabular}{|l|ccccccc|}
\hline 
\rowcolor[gray]{.9} \bf Lang. & \bf LTB & \bf MPB & \bf NEA & \bf NEB & \bf OLB & \bf STB & \bf SUX \\ \hline
\bf LTB & \textit{\textbf{947}} & 6 & 9 & 34 & 13 & 25 & 8 \\
\rowcolor[gray]{.9}\bf MPB & 3 & \textit{\textbf{858}} & 51 & 94 & 84 & 69 & 55 \\
\bf NEA & 6 & 26 & \textit{\textbf{780}} & 185 & 26 & 148 & 26 \\
\rowcolor[gray]{.9}\bf NEB & 4 & 19 & 81 & \textit{\textbf{535}} & 30 & 160 & 30 \\
\bf OLB & 3 & 22 & 12 & 16 & \textit{\textbf{736}} & 47 & 110 \\
\rowcolor[gray]{.9}\bf STB & 17 & 35 & 30 & 113 & 43 & \textit{\textbf{491}} & 101 \\
\bf SUX & 5 & 19 & 22 & 8 & 53 & 45 & \textit{\textbf{655}} \\
\hline
\end{tabular}
\end{center}
\caption{\label{confusion} Confusion matrix for the product of relative frequencies method. The rows indicate the actual languages and the columns indicate predicted languages. Correct identifications are emphasized.}
\end{table}

\section{Conclusions and future work}

In this paper, we have shown that it is possible to perform language and dialect identification in cuneiform texts encoded in Unicode characters. We have created a dataset to be used in the VarDial Evaluation campaign and evaluated the performance of four baseline identifiers using the dataset.

Some sizeable Oracc projects were left out of the corpus, for example the ``Royal Inscriptions of the Neo-Assyrian Period'' project, as the exact dialect of the Akkadian language had not been annotated. Furthermore, for the same reason, only the lines in Sumerian could be used from the ``Royal Inscriptions of Babylonia online (RiBo)'' project. We believe that automatic dialect identification could be useful in making the annotations more detailed and are planning to provide this kind of automatically deduced information as part of the Korp version of Oracc.\footnote{http://urn.fi/urn:nbn:fi:lb-2018071121}

Some other avenues for further work are language set identification for the multilingual texts, as well as unsupervised clustering of data without any predefined languages.

\section*{Acknowledgments}

The research was carried out in the context of the ``Semantic Domains in Akkadian Texts'' project and the Centre of Excellence in Ancient Near Eastern Empires at the University of Helsinki (ANEE), both funded by the Academy of Finland. This work has also been partly funded by the Kone Foundation.

We thank Johannes Bach, Mikko Luukko, Aleksi Sahala, and Niek Veldhuis for their valuable comments during the experiments and Raija Mattila and Saana Sv\"{a}rd for their continued support. We are grateful to Robert Whiting for revising the language of the text and for further insight into the Akkadian and Sumerian languages.

\bibliography{naaclhlt2019}
\bibliographystyle{acl_natbib}

\end{document}